\documentclass[pmlr]{jmlr}

\RequirePackage{graphicx}
 \usepackage{booktabs}
\usepackage{longtable}
  \makeatletter                                                                                                                                   
  \def\set@curr@file#1{\def\@curr@file{#1}} 
  \renewcommand{\@titlefoot}{}                                                                                                                      
  \makeatother   
\usepackage[load-configurations=version-1]{siunitx} 

\theorembodyfont{\upshape}
\theoremheaderfont{\scshape}
\theorempostheader{:}
\theoremsep{\newline}

\jmlrpages{}  
\firstpageno{1}

\title[Agent Memory and Reconciliation]{Detecting Clinical Discrepancies in Health Coaching Agents: A Dual-Stream Memory and Reconciliation Architecture}

\author{\Name{Samuel L Pugh}
       \Email{samuelpugh@verily.health}\\ 
       \addr
       Verily Health Inc.
       \AND
       \Name{Eric Yang}
       \Email{eryang@verily.health}\\ 
       \addr Verily Health Inc.
       \AND
       \Name{Alexander Muir Sutherland}
       \Email{asutherland@verily.health}\\ 
       \addr Verily Health Inc.
       \AND
       \Name{Alessandra Breschi}
       \Email{abreschi@verily.health}\\ 
       \addr Verily Health Inc.
       }

\begin{document}

\maketitle

 \begin{abstract}
 As Large Language Model (LLM) agents transition from single-session tools to persistent systems managing longitudinal healthcare journeys, their memory architectures face a critical challenge: reconciling two imperfect sources of truth. The patient's evolving self-report is current but prone to recall bias, while the Electronic Health Record (EHR) is medically validated but frequently stale. General-purpose agent memory systems optimize for coherence by overwriting older facts with the user's latest statement, a pattern that risks safety failures when applied to clinical data. We introduce a Dual-Stream Memory Architecture that strictly separates the patient narrative from the structured clinical record (FHIR), governed by a dedicated Reconciliation Engine that evaluates every extracted memory against the patient's FHIR profile and classifies discrepancies by type, severity, and the specific FHIR resources involved. We evaluate this architecture on 26 patients across 675 longitudinal wellness coaching sessions, using a hybrid dataset that interleaves real provider-patient transcripts with synthetic, FHIR-grounded clinical scenarios. In isolated testing, the engine detects 84.4\% of designed clinical discrepancies with 86.7\% safety-critical recall. By coupling extraction and reconciliation evaluation on the same data, we directly quantify a 13.6\% error cascade, tracing the degradation to clinical details lost during memory extraction from unstructured conversation rather than to downstream classification errors. These findings establish that validating patient-reported memories against clinical records is both feasible and necessary for safe deployment of longitudinal health agents.
\end{abstract}

\section{Introduction}

Conversational AI driven by Large Language Models (LLMs) has transformed digital health, shifting from static patient portals to highly interactive, scalable interfaces capable of continuous patient engagement, triage, and health education (\cite{Cevasco2024-xm, Mahajan2025-rx, Rashidian2025-ue, Yang2025-us}). However, as these systems transition from stateless, single-session query engines to persistent agents managing long-horizon longitudinal care journeys, they encounter severe architectural shortcomings. The fundamental issue lies not in language generation, but in memory architecture and epistemic uncertainty (\cite{Brender2025-hj}).

Managing patient health information requires reconciling two imperfect sources of truth: the patient and the electronic health record (EHR) (\cite{Christof2025-he}). The patient possesses current, lived experience but is prone to recall bias (\cite{Zini2021-dn}). Conversely, the EHR contains medically validated data but is frequently stale or incomplete (\cite{Gurupur2025-cc}). Current AI architectures exacerbate this dilemma. While prior work on agent memory systems has improved conversational coherence and personalization, healthcare requires strict data integrity. Standard Retrieval-Augmented Generation (RAG) systems flatten these data sources into a single vector space, treating all retrieved context as equally authoritative. Without reasoning about the decay of clinical data, these systems cannot effectively weigh a divergent patient narrative against an aging medical record (\cite{Muller2025-sq}). This architectural flaw inevitably leads to either hallucinated compliance, where the agent accepts an inaccurate self-report, or protocol rigidity, where it rigidly enforces an outdated system record.

Traditionally, healthcare systems mitigate these divergent narratives through reconciliation, an episodic, clinician-initiated workflow typically reserved for structured care transitions (\cite{barnsteiner2008medication}). However, this manual process is ill-equipped to capture the gradual drift in patient behavior and health status that emerges over weeks or months of casual conversation. Digital health agents, such as wellness coaches and chronic disease management tools, interact with patients far more frequently than traditional clinic visits. This high-frequency engagement creates a significant opportunity. Rather than waiting for the next scheduled appointment, these agents can continuously detect EHR staleness and actively reconcile discrepancies in the background of routine conversation.

 To address this gap, we propose a Dual-Stream Memory Architecture for conversational health agents. While persistent memory is essential for tracking patient goals and preferences over longitudinal care journeys, writing unvalidated self-reports directly into an agent's memory risks compounding clinical errors. Our architecture solves this by strictly isolating the patient's evolving narrative from the read-only, structured clinical record (FHIR). Furthermore, we introduce a dedicated Reconciliation Engine that evaluates every new narrative memory against the patient's FHIR clinical record. Rather than silently overwriting facts, the engine actively flags conflicts, such as a patient discontinuing an active medication, and produces a discrepancy classification that traces each finding back to specific FHIR resources. This fundamentally decouples objective discrepancy detection from subjective dialogue generation. 
 
 Our core technical contributions are threefold. First, we implement the dual-stream memory and reconciliation architecture, a modular framework that maintains clinically consistent patient states by continuously validating extracted narrative memories against authoritative clinical records. Second, we introduce a novel hybrid evaluation dataset that seamlessly interleaves authentic wellness coaching transcripts with synthetic, FHIR-grounded clinical scenarios, enabling rigorous longitudinal stress-testing across 675 continuous sessions. Finally, by evaluating extraction and reconciliation both in isolation and sequentially, we provide an end-to-end error cascade analysis that directly quantifies how upstream memory extraction errors degrade downstream clinical detection, revealing critical bottlenecks in multi-stage clinical AI pipelines.

\subsection*{Generalizable Insights about Machine Learning in the Context of Healthcare}

This work offers three insights relevant to the broader ML-for-health community:

  \begin{itemize}
  \item \textbf{Agent memory systems for healthcare must validate patient statements against clinical records.} General-purpose memory systems treat the user's latest statement as ground truth. In healthcare, this paradigm is fundamentally unsafe; a patient reporting they stopped taking a medication should trigger a structured clinical evaluation, not a silent database update. We demonstrate that separating patient-reported memories from the clinical record, and reconciling between them at extraction time, enables reliable discrepancy detection (84.4\%) while maintaining high faithfulness (4.8/5) in the stored memories.

  \item \textbf{Multi-stage clinical AI pipelines require component-level evaluation to identify bottlenecks.} End-to-end evaluation of chained LLM systems can mask where failures originate. By evaluating memory extraction and reconciliation both in isolation and together, we directly measure a 13.6\% error cascade and trace it to the extraction stage, demonstrating that end-to-end evaluation alone would mask the true source of performance degradation.

  \item \textbf{Clinical reconciliation can operate as a modular background process on existing agents.} Rather than building dedicated medical interview agents, we show that EHR reconciliation can operate effectively as a modular, ambient background process layered onto routine, non-clinical interactions (wellness coaching, chronic disease management, patient navigation). This architecture provides a blueprint for layering robust clinical safety guardrails onto existing, proprietary, or domain-specific conversational models.
  \end{itemize}

\section{Related Work}

\subsection{Agent Memory Systems and Benchmarks}
Early cognitive frameworks for AI systems, such as Generative Agents, simulate human recall by partitioning memory into episodic, semantic, and procedural groups (\cite{Park2023-ACM}). Operating system-inspired hierarchies, such as MemGPT (Letta), manage finite context windows and grant agents autonomy to retrieve information via tool calls (\cite{Packer2023-gk}). Production systems like Mem0 layer graph-based stores over vector databases to execute explicit structural operations based on semantic proximity (\cite{Chhikara2025-fb}). Temporally-aware architectures like Zep advance further by maintaining bi-temporal knowledge graphs (\cite{Rasmussen2025-zep}). These systems optimize for conversational coherence, persona alignment, and user satisfaction. When faced with contradictory information, they edit older facts to match the user's latest input. While acceptable in general domains, overwriting medical information based on casual statements degrades the factual integrity of clinical records. 

This same limitation extends to how these systems are evaluated. Benchmarks like LoCoMo, LongMemEval, and MemoryBench focus heavily on an agent's explicit recall capacity rather than its ability to handle contradictions (\cite{Maharana2024-uc, Wu2024-mo, Ai2025-kb}). While recent benchmarks recognize conflict resolution as a critical competency, they fundamentally assume the user narrative as a single source of truth. For instance, MemoryAgentBench (\cite{Hu2025-mab}) demonstrates that frontier models struggle with basic memory conflicts, while frameworks like BeliefShift (\cite{Myakala2026-bs}) measure an agent's susceptibility to temporal opinion drift. However, these evaluations are confined to domains where the user's internal narrative is paramount. Consequently, a critical gap remains: no existing benchmark evaluates the epistemic complexity of healthcare, where an agent must actively detect and reconcile contradictions between a subjective, continuous patient dialogue and an authoritative external database like the EHR.

\subsection{Clinical Agents and the Illusion of Ground Truth}
While general-domain agents prioritize user alignment, healthcare agents swing to the opposite extreme by treating the EHR as infallible.
Specialized benchmarks  (e.g., FHIR-AgentBench, MedAgentBench, FHIR-AgentEval)  and clinical NLP tools (e.g., LLMonFHIR) evaluate an agent's ability to accurately navigate, query, and alter complex patient records (\cite{Lee2025-tu, Jiang-2025-mab, Mokssit2026-fj, Schmiedmayer2025-qp, Qu2026-ky, Alsentzer2019-no, Rasmy2021-yk}). By defining success purely on adherence to the provided FHIR environment, these prior works ignore the temporal staleness and fragmentation inherent in real-world EHRs, optimizing agents for a sanitized reality. This leaves them highly vulnerable to epistemic uncertainty when a patient's actual health status evolves. In contrast, our architecture explicitly rejects the assumption of EHR infallibility. We treat the clinical record not as static ground truth, but as a historical baseline that must be continuously validated against the patient's lived experience.

\subsection{Conversational Medication Reconciliation}
To address patient information drift, systems like AMREC actively probe patients for medication discordances (\cite{Deo2025-lk}). However, because they function as dedicated medical interviewers, they rely on structured interrogation. This makes them poorly equipped for casual, unstructured interactions where clinical information emerges organically. Similarly, LLMs have been evaluated on static medication reconciliation tasks using structured prompts rather than natural dialogue (\cite{Sridharan2024-medrec}).   Rather than treating reconciliation as an explicit dialogue task, our system embeds it passively within the background memory pipeline. By extracting memories from natural conversations and flagging discrepancies against the FHIR profile, our architecture can route actionable alerts without interrupting the natural conversation flow.

\section{Methods}

\subsection{Problem Formulation}
We frame clinical-narrative reconciliation as a continuous process of tracking patient state and comparing it across two distinct data streams. At dialogue turn $t$, given a history $H_t=(u_1, a_1, \dots, a_{t-1}, u_t)$ of patient utterances $u$ and agent responses $a$, the system evaluates extracted narrative updates against a static clinical environment. To handle the uncertainty of conflicting information, we separate the data into two independent streams. The Narrative Stream ($M_N$) captures patient-reported assertions, defined prior to turn $t$ as $M_N^{(t-1)}=\{m_1, \dots, m_k\}$, where each memory $m_i$ represents a single piece of patient-reported information (e.g., a health status, medication, or preference). The Clinical Stream ($M_C$) is a read-only FHIR R4 bundle defined as $M_C=\{e_1, \dots, e_n\}$. Each clinical entity $e_j \in M_C$ is a tuple $e_j=(v_j, \lambda_j, \rho_j)$ containing the clinical value $v_j$ (e.g., an active diagnosis), the resource ontology $\lambda_j$ (e.g., Condition), and temporal metadata $\rho_j$ to gauge data staleness. Instead of passively appending text to a context window, an LLM-based extraction function $f_{ext}$ identifies what changed in each turn, producing explicit delta-operations:$$\Delta M_N^{(t)}=f_{ext}(u_t, M_N^{(t-1)})$$where $\Delta M_N^{(t)}$ is the subset of $\mathcal{O} \in \{\text{insert}, \text{update}, \text{delete}\}$ operations generated at turn $t$. By processing only these updates, the Reconciliation Engine $f_{rec}$ compares each altered memory $m_i \in \Delta M_N^{(t)}$ against the clinical stream $M_C$ to identify conflicts or gaps. It outputs a discrepancy characterization tuple:$$R_i=f_{rec}(m_i, M_C)=(d, E_c, s, \sigma, \psi)$$Here, $d \in \{0, 1\}$ indicates whether a clinically relevant discrepancy was detected (contradiction or information gap), $E_c \subseteq M_C$ identifies the FHIR resources involved, and $s \in \{\text{low, medium, high}\}$ denotes clinical significance. The boolean $\sigma \in \{0, 1\}$ indicates whether the discrepancy poses immediate patient safety risk (e.g., a contraindicated medication or unreported allergy), enabling downstream systems to prioritize urgent findings. $\psi$ provides the model's natural language justification. This formalization strictly grounds reconciliation in traceable FHIR resources.

\subsection{Dual-Stream Memory Architecture}

\begin{figure}[t]
   \centering 
   \includegraphics[width=1.0\textwidth]{figures/system_diagram.png} 
   \caption{Dual-Stream memory architecture featuring narrative and clinical streams, a delta-based extraction system, and a reconciliation engine for discrepancy characterization and downstream action.}
   \label{fig:example} 
 \end{figure} 

 To implement the reconciliation formalized above, we introduce a dual-stream architecture that isolates patient-reported memories from structured clinical data (Figure \ref{fig:example}).

\subsubsection{The Narrative Stream and Delta-Based Extraction}
The narrative stream captures patient information via free-form content, categorical tags, and strict temporal metadata. To prevent memory fragmentation, the system separates information into two distinct types: stable facts (which receive persistent records) and evolving states (which are updated in-place to preserve their temporal trajectory). During extraction, an LLM pipeline processes the raw dialogue to generate structured \texttt{MemoryDelta} updates (Appendix~\ref{app:memory_schema}). Rather than rewriting the entire memory state from scratch, the model extracts only what has changed, outputting discrete \texttt{insert}, \texttt{update}, and \texttt{delete} operations. Duplicate prevention is rigorously enforced through prompt instructions, exact-match guards, and inline ID routing (see Appendix~\ref{app:extraction_prompts} for the full extraction prompt). Finally, the system maintains two memory views: an extraction view containing IDs and metadata for precise LLM editing, and a clean, categorized view for the downstream agent or practitioner.

\subsubsection{The Clinical Stream}The clinical stream ($M_C$) represents the patient's objective medical history, constructed from a curated FHIR R4 bundle. To optimize the context window, the clinical summary is curated to approximately 1100-1600 tokens, retaining active or resolved conditions, medications,  allergies, and the most recent values for relevant observations (labs, vitals, procedures, encounters). All resources include structured inline tags formatted as \texttt{[ResourceType] display [SYSTEM:CODE]} (e.g., \texttt{[SNOMED-CT:59621000]}). This explicit tagging enables the Reconciliation Engine to output precise \texttt{FHIRResourceRef} citations, ensuring rigorous resource-level precision and recall matching during evaluation (see Appendix~\ref{app:clinical_summary} for a sample clinical summary).

\subsection{The Reconciliation Engine}
To evaluate conflicts, the Reconciliation Engine leverages LLM-driven temporal reasoning. The model is explicitly prompted to contextualize the temporal metadata ($\rho_j$) of each FHIR resource. Through this qualitative reasoning process, the LLM assesses that stale records (e.g., a medication list from two years ago) warrant lower clinical confidence, whereas persistent historical constraints (e.g., allergies) retain high authority regardless of their age. To ensure reproducible measurement, the engine translates this prompt-driven reasoning into an Observable Discrepancy Framework via structured JSON outputs. The engine classifies every narrative update into one of four states:                                                                       
  \begin{itemize}
      \item \textbf{Agreement}: the patient's statement is consistent with the clinical record.
      \item \textbf{Contradiction}: the statement directly conflicts with documented clinical data.
      \item \textbf{Gap}: the patient reports clinically actionable information that is absent from the record (e.g., an undocumented medication,   unreported symptom, or condition the patient reports as resolved).
      \item \textbf{No clinical relevance}: routine lifestyle content (step goals, preferences, daily routines) with no bearing on the FHIR record.        
  \end{itemize}

If an actionable discrepancy is detected, the engine characterizes the discrepancy along several dimensions. Clinical significance ($s$) provides an ordinal classification (low, medium, high) reflecting potential for patient harm, ranging from limited clinical risk (e.g., a condition reported as resolved) to direct safety concerns (e.g., medication self-discontinuation). The safety criticality flag ($\sigma$) is a strict boolean indicating whether the discrepancy poses immediate patient safety risk. This flag is orthogonal to severity. For instance, a discrepancy with moderate clinical magnitude, such as an ambiguous medication adherence pattern (e.g., taking half the prescribed dose) or a risky behavioral change (e.g., starting high-intensity training with uncontrolled hypertension), can still be flagged as safety-critical if failing to address it could lead to acute harm. Both labels are assigned during scenario generation rather than by clinical adjudication, a limitation discussed in Section \ref{sec:discussion}. Furthermore, Precise FHIR Mapping ($E_c$) outputs structured FHIR Resource Ref citations containing the resource type, ontology, and value (Appendix~\ref{app:recon_schema}).  This code-level matching ensures that detected discrepancies are traceable to specific clinical records (see Appendices~\ref{app:recon_prompt} and~\ref{app:recon_result} for the full prompt and a sample output).

\subsection{Modular Integration}
The dual-stream memory and reconciliation engine is designed as a modular layer that can sit on top of any conversational agent. At each patient turn, the extraction module produces narrative deltas and the reconciliation engine evaluates them against the clinical stream, producing structured discrepancy outputs. The engine operates on extracted memories rather than raw conversation text, ensuring its inputs are self-contained semantic units that do not require conversational context for disambiguation. How these outputs are acted upon (e.g., provider alerts, conversational clarification, or safety stops) is left to the deployment context and is not evaluated in this work. Here, we evaluate the extraction and reconciliation components via transcript replay, isolating their performance from both dialogue generation and downstream action handling.

\section{Experimental Setup}

We utilize GPT-4o for delta-based memory extraction and discrepancy classification, selected for its strong structured output capabilities and consistent adherence to schema constraints \cite{OpenAI2024-ac}. Synthetic data generation (scenario creation, hybrid transcript generation, and ground-truth memory extraction) was performed using Gemini 2.5 Pro, chosen for its large context window which accommodates full FHIR bundles alongside conversation history in a single call. Automated evaluation judgments use GPT-4o for ground-truth memory matching (requiring precise semantic comparison) and GPT-4.1 for transcript-level quality assessment (requiring comprehension of full multi-session transcripts).

\subsection{Dataset Composition}
Our evaluation is grounded in authentic patient interactions by modeling a longitudinal wellness coaching agent designed to help patients set step goals, track physical activity, and discuss behavioral barriers. The foundational narrative data driving this simulation is sourced from the Virtual Coach dataset developed by the University of Illinois Chicago Natural Language Processing (UIC NLP) Lab, which contains real, de-identified text-message dialogues between patients and human health coaches (\cite{gupta-etal-2020-human}). Utilizing this corpus, we deterministically segmented 3,665 dialogue turns across 26 patients into 432 discrete, multi-turn wellness coaching sessions using temporal heuristics: messages within a 4-hour window were grouped into the same session, with a maximum session span of 48 hours (average 8.5 turns per session). While highly realistic, these authentic transcripts primarily focus on lifestyle habits and behavioral motivation, naturally lacking the clinical contradictions required to rigorously stress-test a medical reconciliation engine.

Because comprehensive system evaluation demands a robust clinical environment, we inferred clinical demographic profiles from the conversational dataset and passed these parameters into Synthea to generate 50 to 100 candidate FHIR R4 bundles per patient (\cite{Walonoski2018-ci}). These candidates were evaluated using a multi-factor scoring model prioritizing condition and medication overlap, resulting in 26 highly plausible, temporally aligned FHIR bundles, one per patient. Each selected bundle represents a comprehensive longitudinal clinical history, averaging 26 unique conditions (e.g., hypertension, prediabetes, chronic pain), 7 distinct medications, 47 documented procedures, and 3 immunization records per patient, consistent with the multi-morbidity profiles typical of the UIC coaching population. Synthea parameters were configured per patient using inferred age, gender, and target condition modules derived from the conversational data. Using each patient's FHIR profile and UIC transcript as context, we generated 243 synthetic reconciliation scenarios (9.3 per patient on average), of which 83 (34\%) are designated safety-critical. The scenarios reference 390 individual FHIR resources across seven resource types (see Table~\ref{tab:fhir_types} for the per-type breakdown), with Conditions (123 references) and MedicationRequests (109) being the most frequently tested. Severity is distributed across low (119 scenarios), medium (71), and high (53). Crucially, all scenarios maintain the coaching scope constraint: clinical information is introduced not via explicit medical questionnaires, but organically as patients discuss physical barriers or context for their activity goals.

\subsection{Hybrid-Real Simulation Strategy}
\begin{figure}[t]
     \centering
     \includegraphics[width=1.0\textwidth]{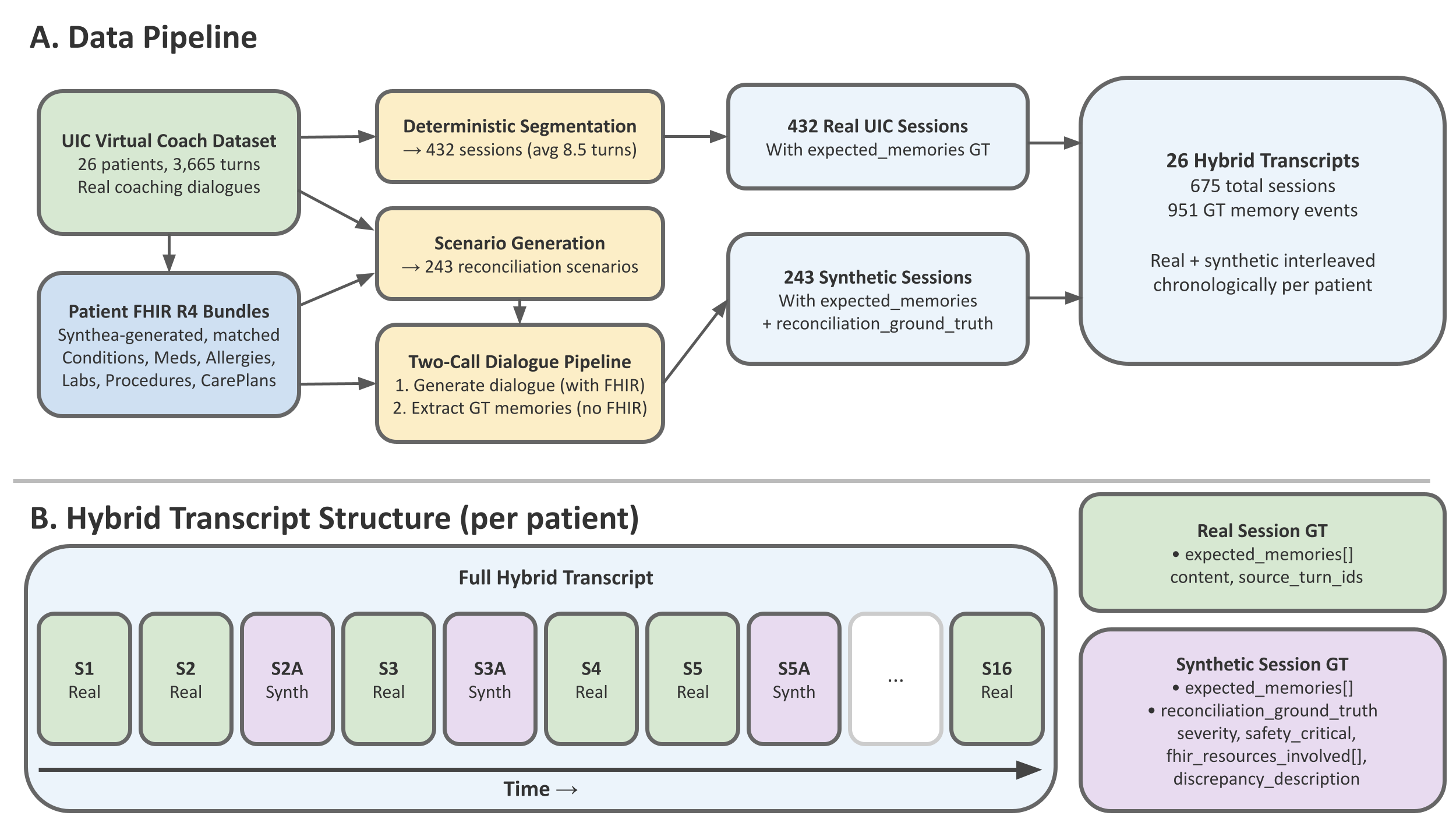}
     \caption{Hybrid transcript construction pipeline. (A) Real UIC coaching transcripts are deterministically segmented into 432        
  sessions, while patient-matched FHIR R4 bundles are used to generate 243 synthetic reconciliation scenarios via a two-call pipeline    
  that separates dialogue generation (with FHIR access) from ground-truth memory extraction (transcript only). (B) Real and synthetic    
  sessions are interleaved chronologically per patient, producing 26 hybrid transcripts with 675 total sessions and 951 ground-truth     
  memory events. Real sessions carry expected memory annotations; synthetic sessions additionally carry reconciliation ground truth    
  (severity, safety flag, FHIR resource references).}
     \label{fig:hybrid_pipeline}
  \end{figure}

Maintaining conversational cohesiveness is paramount when evaluating a longitudinal memory system. Rather than testing the reconciliation engine on isolated clinical vignettes, we sought to simulate a realistic coaching trajectory where clinical issues emerge naturally amidst baseline wellness discussions. We achieved this by interleaving the full conversational dialogues derived from the synthetic clinical scenarios directly into the chronological sequence of the real UIC coaching sessions (See Figure~\ref{fig:hybrid_pipeline}). The resulting dataset comprises 26 unified hybrid transcripts totaling 675 sessions (432 real UIC sessions plus 243 synthetic sessions) and yielding 951 ground-truth memory events. Integrating authentic conversational dynamics with controlled clinical injections provides the continuous longitudinal evaluation data necessary for rigorous end-to-end testing.

Generating conversational dialogues and ground truth memory references for these synthetic scenarios required a two-call pipeline (Figure~\ref{fig:hybrid_pipeline}). During the first step, a dialogue generation model utilized full access to the patient's FHIR context to write the clinical scenario's back-and-forth conversation, mirroring conversational style of the real UIC sessions. Subsequently, a ground-truth memory extraction step processed only this generated conversational transcript, deliberately isolated from any underlying FHIR access. This extraction process was applied to both the synthetic sessions and the real UIC coaching sessions, ensuring every transcript carried expected memory annotations to serve as a unified ground truth across the 675 sessions.

\subsection{Multi-Dimensional Evaluation Framework}
Evaluating complex, multi-stage pipelines requires isolating individual component performance before assessing the system holistically. Replaying fixed conversational transcripts rather than simulating live, dynamic agent responses guarantees  reproducibility and insulates our metrics from downstream dialogue generation variance. We therefore structure our evaluation across three distinct dimensions, allowing us to pinpoint exact failure modes within the architecture before measuring the compounding effects of system-wide integration.

\textbf{Dimension 1: Memory Extraction Quality.} The first dimension measures the accuracy of the delta-based memory extraction pipeline across the full hybrid transcript corpus (675 sessions, 2{,}296 patient turns). The evaluation targets 951 ground-truth memory events: 522 from real UIC sessions (1.2 per session on average) and 429 from synthetic reconciliation sessions (1.8 per session). For each ground-truth event, an LLM judge evaluates whether the expected information is captured in the system's memory state at the corresponding point in the conversation. The judge considers information spread across multiple memories, classifying each event as \textsc{match} (fully captured), \textsc{partial} (gist present but key details missing), or \textsc{no\_match} (not captured). We report \emph{recall} (match + partial / total) and \emph{strict recall} (match only / total) as primary metrics. A separate transcript-level LLM judge assesses the final memory state against the complete conversation for \emph{faithfulness} (absence of hallucinated content) and \emph{deduplication} (absence of redundant memories), each scored 1--5 (see Appendix~\ref{app:judge_prompts} for both judge prompts).
               
\textbf{Dimension 2: Isolated Reconciliation.} The second dimension assesses the reconciliation engine in isolation, independent of upstream extraction errors. Ground-truth narrative memories are fed directly into the classification module, and performance is measured against the 243 synthetic reconciliation scenarios. Each scenario carries a ground-truth label specifying the expected discrepancy, its clinical severity (low, medium, or high), a safety-critical flag, and the specific FHIR resources involved. We evaluate five metrics: (1) \emph{binary detection}: whether the engine flagged any discrepancy in the session; (2) \emph{resource-informed detection}: whether the engine flagged a discrepancy and cited at least one of the expected FHIR resources, measured via exact \texttt{(resource\_type, code\_system, code\_value)} tuple matching; (3) \emph{safety recall}: the fraction of safety-critical scenarios detected; (4) \emph{severity accuracy}: both exact match and within-one-level agreement with ground-truth severity; and (5) \emph{resource recall and precision}: the overlap between engine-cited and ground-truth FHIR resource codes. To characterize the engine's behavior on naturalistic data, we also process memories extracted from the 432 real UIC sessions, which contain no designed discrepancies, and report the distribution of classification types.

\textbf{Dimension 3: End-to-End Pipeline and Error Cascade.} The final dimension evaluates the complete pipeline by processing the hybrid transcripts through extraction and then into reconciliation. The reconciliation stage consumes the \emph{exact same} extracted memories produced during Dimension~1, coupling the two stages and ensuring the error cascade analysis reflects the actual impact of extraction quality on downstream detection. Comparing detection accuracy between Dimensions~2 and~3 quantifies how upstream extraction failures (e.g., lost clinical details, over-consolidated memories) degrade downstream discrepancy detection.

\section{Results}

We evaluated 26 patients across 675 sessions (432 real UIC + 243 synthetic reconciliation), comprising 2{,}296 patient turns and 951 ground-truth memory events.

\subsection{Dimension 1: Memory Extraction}

\begin{table}[t]
  \centering
  \caption{Memory extraction quality across 26 patients (675 hybrid sessions, 951 ground-truth events). Recall measures the fraction of GT events captured fully or partially in the memory state. Faithfulness and deduplication are scored 1--5 by an LLM transcript judge.}
  \begin{tabular}{lcc}
  \toprule
    \textbf{Metric} & \textbf{Mean [95\% CI]} & \textbf{Range} \\
    \midrule
    GT Recall (match + partial) & 0.708 [0.674, 0.740] & 0.49--0.88 \\
    Strict Recall (match only) & 0.262 [0.221, 0.305] & 0.07--0.53 \\
    Faithfulness (/5) & 4.8 [4.62, 4.92] & 4--5 \\
    Deduplication (/5) & 3.6 [3.35, 3.92] & 3--5 \\
    \midrule
    Patient turns processed & \multicolumn{2}{c}{2{,}296} \\
    Final memories extracted & \multicolumn{2}{c}{408} \\
  \bottomrule
  \end{tabular}
  \label{tab:extraction}
\end{table}

Table~\ref{tab:extraction} summarizes extraction performance. The system captures 70.8\% of ground-truth events (25.2\% full match, 44.7\% partial, 30.1\% no match). Partial matches typically reflect the consolidation architecture working as designed: the system maintains the current state of evolving information (e.g., the latest step goal) rather than preserving every historical detail. Faithfulness scores are high (4.8/5), indicating that the system rarely fabricates information not present in the conversation. Deduplication (3.6/5) is the weakest dimension, with goal-tracking conversations prone to redundant memory entries when patients express similar intentions across sessions.

  \subsubsection{Human Expert Validation of Extracted Memories}
  Two domain experts independently reviewed transcripts and rated all 104 GT memory events across 4 patients (80 sessions) for accuracy and clinical relevance using the rubric in Appendix~\ref{app:human_annotation}. Neither annotator rated any event as Inaccurate, and both agreed on at least Partially Accurate for 100\% of events and at least Useful for 99\% of events. Annotators showed 100\% within-one-level agreement on both dimensions, with Gwet's AC1 (\cite{Gwet2008-ac1}) indicating substantial agreement (0.80 for accuracy, 0.67 for relevance). Review of annotator-flagged missing information identified only 6 instances across 80 sessions where information within the system's intended scope was not extracted. These results validate the quality of the GT memory events used throughout our evaluation.

\subsection{Dimension 2: Isolated Reconciliation}

\begin{table}[t]
  \centering
  \caption{Reconciliation performance on 243 synthetic scenarios using ground-truth memories (Dimension~2) versus extracted memories from the full pipeline (Dimension~3). The delta column quantifies the error cascade from upstream extraction. 95\% bootstrap confidence intervals in brackets.}
  \begin{tabular}{lccc}
  \toprule
    \textbf{Metric} & \textbf{Isolated (GT Memories)} & \textbf{Full Pipeline} & \textbf{$\Delta$} \\
    \midrule
    Detection rate & 84.4\% [79.8, 88.9] & 70.8\% [65.0, 76.5] & $-$13.6\% \\ 
    Resource-informed detection & 76.5\% [71.2, 81.9] & 62.6\% [56.4, 68.7] & $-$14.0\% \\
    Safety recall & 86.7\% [79.5, 94.0] & 75.9\% [66.3, 84.3] & $-$10.8\% \\
    Severity within-1 & 81.5\% [76.5, 86.4] & 67.1\% [61.3, 72.8] & $-$14.4\% \\
    Severity exact match & 28.0\% [22.6, 33.7] & 25.5\% [20.2, 30.9] & $-$2.5\% \\
    Resource recall (mean) & 65.3\% [60.0, 70.4] & 51.4\% [45.9, 56.9] & $-$13.9\% \\
    Resource precision (mean) & 71.2\% [66.3, 76.1] & 62.1\% [56.5, 67.7] & $-$9.1\% \\
  \bottomrule
  \end{tabular}
  \label{tab:reconciliation}
\end{table}

Table~\ref{tab:reconciliation} presents reconciliation accuracy. In isolated testing (Dimension~2), the engine detects 84.4\% of designed discrepancies with 86.7\% safety recall. We report two detection metrics: \emph{binary detection} (did the engine flag any discrepancy in the session?) and \emph{resource-informed detection} (did it cite at least one of the expected FHIR resources?). The 7.9\% gap between these metrics (84.4\% vs.\ 76.5\%) reflects cases where the engine identified the correct clinical concern but cited a related rather than identical FHIR resource, for example citing the underlying condition rather than the specific medication. Review of these 19 cases confirmed that each identified the correct clinical concern but cited a related resource (e.g., a broader condition code rather than a specific procedure, or a diagnosis rather than the associated care plan). Of the 205 detected discrepancies in Dimension~2, 29 (14.1\%) were classified as contradictions and 176 (85.9\%) as information gaps, consistent with coaching conversations surfacing undocumented patient information more frequently than direct conflicts with the clinical record.

Severity classification achieves 81.5\% within-one-level accuracy but only 28.0\% exact match. The engine tends to underestimate severity for complex, multi-factor scenarios (e.g., a patient planning high-intensity exercise with concurrent hypertension and a recent procedure) while occasionally overestimating low-severity information gaps. Detection rates were consistent across severity levels (84.0\% low, 83.1\% medium, 86.8\% high), with all levels experiencing a similar 13--14\% 
  cascade loss through the full pipeline. 

\subsection{Dimension 3: Error Cascade}

The coupled pipeline reveals a 13.6\% absolute reduction in detection rate from Dimension~2 to Dimension~3 (84.4\% $\rightarrow$ 70.8\%). Of the 39 detections lost through the cascade, 54\% involved low-severity scenarios, 28\% medium, and 18\% high-severity. Ten safety-critical scenarios were lost, in each case because the memory extraction system failed to capture the relevant clinical detail, for example consolidating a medication adherence statement into a broader lifestyle memory that lost the specific drug reference. Six scenarios were gained in Dimension~3 (detected in the full pipeline but missed in isolation), in cases where the extraction system produced memories that framed the clinical content differently than the ground truth, occasionally triggering detection where the original GT phrasing did not. The 13.6\% detection degradation is statistically significant (bootstrap 95\% CI for the paired difference: [8.6\%, 18.9\%]). Table~\ref{tab:qualitative} in Appendix~\ref{app:qualitative} illustrates representative success and failure cases from the pipeline.

\subsection{Effect of Conversation Length}

Extraction recall shows a moderate negative correlation with the number of patient turns ($r = -0.46$, $p = 0.018$) and with the final memory count ($r = -0.48$, $p = 0.013$). This reflects the increasing difficulty of maintaining comprehensive memory state as the consolidation system merges and updates memories over many sessions. In contrast, reconciliation detection rate shows no significant correlation with conversation length (full pipeline detection rate vs. patient turns: $r = 0.12$, $p = 0.56$; vs. final memory count: $r = 0.02$, $p = 0.93$), confirming that the engine performs consistently regardless of accumulated history.  This asymmetry suggests that memory extraction and management, not reconciliation capacity, is the primary bottleneck for longitudinal deployment.

\subsection{Specificity: Real UIC Sessions}

To assess specificity, we examined reconciliation results on the 432 real UIC sessions, which contain authentic coaching conversations without designed clinical discrepancies. Of 1{,}124 total reconciliation results, 83.3\% were classified as no clinical relevance, 16.6\% as information gaps, and fewer than 1\% as \texttt{agreement}. Manual review confirmed these gaps represent genuine clinical content emerging organically in conversation (e.g., medication changes, symptom reports). Zero contradictions were flagged, indicating the engine does not fabricate conflicts where none exist.

\subsection{FHIR Resource Type Coverage}

\begin{table}[t]
  \centering
  \caption{Detection rate by FHIR resource type. All seven resource types in the ground truth are covered,  with AllergyIntolerance and Immunization achieving perfect detection in isolation.}
  \begin{tabular}{lccc}
  \toprule
    \textbf{Resource Type} & \textbf{Scenarios} & \textbf{Isolated} & \textbf{Full Pipeline} \\
    \midrule
      AllergyIntolerance & 6 & 100.0\% & 66.7\% \\
      CarePlan & 60 & 66.7\% & 58.3\% \\
      Condition & 121 & 86.0\% & 74.4\% \\
      Immunization & 25 & 100.0\% & 92.0\% \\
      MedicationRequest & 92 & 89.1\% & 77.2\% \\
      Observation & 28 & 82.1\% & 57.1\% \\
      Procedure & 32 & 75.0\% & 65.6\% \\ 
  \bottomrule
  \end{tabular}
  \label{tab:fhir_types}
\end{table}

Table~\ref{tab:fhir_types} shows detection rates across all seven FHIR resource types. Discrete, fact-based resources (AllergyIntolerance and Immunization at 100\%, MedicationRequest at 89.1\%) achieve the highest isolated detection rates. Contextual resources (CarePlan at 66.7\%, Procedure at 75.0\%) are harder to detect, reflecting the difficulty of identifying conflicts with care plan recommendations and laboratory values in unstructured coaching dialogue.

\subsection{Model Sensitivity}

A natural question is whether the architecture's performance is driven by the specific LLM or by the architectural design itself. To investigate, we repeated the full evaluation using GPT-5.4, a more recent model, in place of GPT-4o for both memory extraction and discrepancy classification. Table~\ref{tab:model_comparison} presents the comparison across all three dimensions.                                               
\begin{table}[t]
      \centering
      \caption{Model comparison across all dimensions. Metrics defined in Tables~\ref{tab:extraction} and~\ref{tab:reconciliation}.}       
      \small
      \begin{tabular}{lcccccccccc}
      \toprule
        & \multicolumn{4}{c}{\textbf{Dim.~1: Extraction}} & \multicolumn{3}{c}{\textbf{Dim.~2: Isolated}} & \multicolumn{3}{c}{\textbf{Dim.~3: Pipeline}} \\        
        \cmidrule(lr){2-5} \cmidrule(lr){6-8} \cmidrule(lr){9-11}
        & Recall & Strict & Faith. & Dedup. & Det. & Safety & Res.-inf. & Det. & Safety & Res.-inf. \\
      \midrule
        GPT-4o & 0.708 & 0.262 & 4.8 & 3.6 & 84.4\% & 86.7\% & 76.5\% & 70.8\% & 75.9\% & 62.6\% \\
        GPT-5.4 & 0.674 & 0.312 & 5.0 & 4.7 & 73.3\% & 80.7\% & 69.5\% & 56.4\% & 66.3\% & 49.8\% \\
      \bottomrule
      \end{tabular}
      \label{tab:model_comparison}
    \end{table}
  
  GPT-4o achieves higher detection (84.4\% vs.\ 73.3\%), safety recall (86.7\% vs.\ 80.7\%), and resource-informed detection (76.5\% vs.\ 69.5\%), while GPT-5.4 produces higher faithfulness (5.0 vs.\ 4.8) and better deduplication (4.7 vs.\ 3.6). This pattern suggests a precision-recall tradeoff: GPT-5.4 is more conservative, defaulting to agreement or no clinical relevance where GPT-4o commits to flagging a potential discrepancy. For a clinical surveillance application where missed discrepancies carry higher risk than false alarms, the more aggressive detection behavior of GPT-4o is preferable. Resource recall is comparable between models (65.3\% vs.\ 66.0\%), indicating that when either model detects a discrepancy, it identifies the relevant FHIR resources with similar accuracy. These results suggest that the architecture's performance is sensitive to model choice, and that newer models do not automatically improve clinical detection. The primary contribution of this work is the architectural framework and evaluation methodology, not optimization of any particular model configuration.

\section{Discussion}
\label{sec:discussion}

\subsection{Principal Findings}

Our work demonstrates that conversational agents can actively reconcile evolving patient narratives with clinical records as a background process. While existing memory systems optimize for coherence by treating the user's latest statement as ground truth, the dual-stream architecture maintains the integrity of both sources by validating extracted memories against a FHIR-grounded clinical profile. The engine detects 84.4\% of designed discrepancies in isolation with 86.7\% safety-critical recall and high faithfulness (4.8/5), showing that automated discrepancy detection is feasible without introducing hallucinated clinical content.

By evaluating extraction and reconciliation both in isolation and end-to-end, we directly quantified a 13.6\% error cascade from memory extraction to discrepancy detection. This analysis reveals that the degradation stems from upstream extraction losing clinical details during memory consolidation, not from downstream reasoning errors. Extraction recall degrades with conversation length ($r = -0.46$, $p = 0.018$) while reconciliation detection does not ($r = 0.12$, $p = 0.56$), directly identifying extraction as the primary bottleneck for longitudinal deployment. These findings highlight that multi-stage clinical AI systems require component-level evaluation to identify where improvements matter most.

\subsection{Clinical Implications}                              
In current practice, medication and record reconciliation is an episodic, clinician-driven task, leaving informational gaps when patients adjust behaviors or experience adverse events between formal visits. By demonstrating that clinical validation can operate continuously in the background of unstructured wellness coaching conversations, this work shows the feasibility of ambient clinical surveillance through routine patient-agent interactions.     

Because the reconciliation engine isolates detection from dialogue generation, it avoids interrupting conversational flow to act as a medical interrogator. Instead, it produces structured, severity-graded outputs mapped to specific FHIR resources, enabling flexible downstream actions: safety-critical findings can be routed directly to care teams, moderate discrepancies can prompt conversational clarification, and low-severity information gaps can be logged for the next scheduled visit. This modularity allows the reconciliation layer to be added to existing digital health applications (wellness coaching, chronic care management, patient navigation) without modifying the primary conversational experience, though determining how best to act on detected discrepancies (e.g., when to alert a clinician versus prompt the patient for clarification) remains an open question for future deployment studies. 

\subsection{Limitations and Future Directions}

To evaluate the pipeline across hundreds of longitudinal sessions, we used Synthea-generated FHIR bundles matched to real patients via fuzzy scoring, and LLM-generated ground-truth labels for severity and safety criticality. While binary detection and resource identification is robust to label quality, severity and safety recall metrics should be interpreted with this caveat. The synthetic reconciliation conversations, while constrained to coaching scope and seeded by real UIC transcripts, may not fully capture the ambiguity of real patient speech. Future work should validate the engine on real patient conversations paired with their actual EHR data and clinician-adjudicated ground-truth labels.

Our analysis also identified that extraction recall degrades with conversation length, indicating that the consolidation architecture struggles to maintain comprehensive state over many sessions. Scaling strategies such as semantic retrieval during consolidation, feeding the extraction module only contextually relevant prior memories rather than the full set, may mitigate this decay but have not been evaluated. More broadly, a prospective deployment study with real patients and real EHR data would test whether the patterns observed here generalize to clinical practice and provide actionable value in routine care workflows.


\bibliography{sample}

@ARTICLE{Mahajan2025-rx,
  title     = "Transforming healthcare delivery with conversational {AI}
               platforms",
  author    = "Mahajan, Arjun and Powell, Dylan",
  journal   = "NPJ Digit. Med.",
  publisher = "Springer Science and Business Media LLC",
  volume    =  8,
  number    =  1,
  pages     = "581",
  month     =  sep,
  year      =  2025,
  copyright = "https://creativecommons.org/licenses/by-nc-nd/4.0",
  language  = "en"
}

@ARTICLE{Cevasco2024-xm,
  title     = "Patient engagement with conversational agents in health
               applications 2016-2022: A systematic review and meta-analysis",
  author    = "Cevasco, Kevin E and Morrison Brown, Rachel E and Woldeselassie,
               Rediet and Kaplan, Seth",
  journal   = "J. Med. Syst.",
  publisher = "Springer Science and Business Media LLC",
  volume    =  48,
  number    =  1,
  pages     = "40",
  month     =  apr,
  year      =  2024,
  copyright = "https://creativecommons.org/licenses/by/4.0",
  language  = "en"
}

@ARTICLE{Rashidian2025-ue,
  title        = "{AI} agents for conversational patient triage: Preliminary
                  simulation-based evaluation with real-world {EHR} data",
  author       = "Rashidian, Sina and Li, Nan and Amar, Jonathan and Lee, Jong
                  Ha and Pugh, Sam and Yang, Eric and Masterson, Geoff and Cha,
                  Myoung and Jia, Yugang and Vaid, Akhil",
  year         =  2025,
  primaryClass = "cs.CL",
  eprint       = "2506.04032"
}

@ARTICLE{Yang2025-us,
  title     = "From barriers to tactics: Development study of behavioral
               science-informed agentic workflow for personalized nutrition
               coaching",
  author    = "Yang, Eric and Garcia, Tomas and Williams, Hannah and Kumar,
               Bhawesh and Ram{\'e}, Martin and Rivera, Eileen and Ma, Yiran
               and Amar, Jonathan and Catalani, Caricia and Jia, Yugang",
  journal   = "JMIR Form. Res.",
  publisher = "JMIR Publications Inc.",
  volume    =  9,
  number    = "v9i8e75421",
  pages     = "e75421",
  month     =  aug,
  year      =  2025,
  language  = "en"
}

@ARTICLE{Brender2025-hj,
  title     = "Clinical notes as narratives: Implications for large language
               models in healthcare",
  author    = "Brender, Teva D and Celi, Leo A and Cobert, Julien M",
  journal   = "J. Gen. Intern. Med.",
  publisher = "Springer Science and Business Media LLC",
  volume    =  40,
  number    =  3,
  pages     = "687--689",
  month     =  feb,
  year      =  2025,
  copyright = "https://www.springernature.com/gp/researchers/text-and-data-mining",
  language  = "en"
}

@ARTICLE{Christof2025-he,
  title     = "Implications of integrating large language models into clinical
               decision making",
  author    = "Christof, Michael and Armoundas, Antonis A",
  journal   = "Commun. Med. (Lond.)",
  publisher = "Springer Science and Business Media LLC",
  volume    =  5,
  number    =  1,
  pages     = "490",
  month     =  nov,
  year      =  2025,
  copyright = "https://creativecommons.org/licenses/by-nc-nd/4.0",
  language  = "en"
}

@ARTICLE{Zini2021-dn,
  title     = "A narrative literature review of bias in collecting patient
               reported outcomes measures ({PROMs})",
  author    = "Zini, Michela Luciana Luisa and Banfi, Giuseppe",
  journal   = "Int. J. Environ. Res. Public Health",
  publisher = "MDPI AG",
  volume    =  18,
  number    =  23,
  pages     = "12445",
  month     =  nov,
  year      =  2021,
  copyright = "https://creativecommons.org/licenses/by/4.0/",
  language  = "en"
}

@ARTICLE{Gurupur2025-cc,
  title     = "Incompleteness of electronic health records: An impending
               process problem within healthcare",
  author    = "Gurupur, Varadraj and Hooshmand, Sahar and Prabhu, Deepa
               Fernandes and Trader, Elizabeth and Salvi, Sanket",
  journal   = "Healthcare (Basel)",
  publisher = "MDPI AG",
  volume    =  13,
  number    =  22,
  pages     = "2900",
  month     =  nov,
  year      =  2025,
  copyright = "https://creativecommons.org/licenses/by/4.0/",
  language  = "en"
}

@ARTICLE{Muller2025-sq,
  title         = "Data quality challenges in {Retrieval-Augmented} Generation",
  author        = "M{\"u}ller, Leopold and Holstein, Joshua and Bause, Sarah
                   and Satzger, Gerhard and K{\"u}hl, Niklas",
  month         =  oct,
  year          =  2025,
  copyright     = "http://creativecommons.org/licenses/by-nc-nd/4.0/",
  archivePrefix = "arXiv",
  primaryClass  = "cs.AI",
  eprint        = "2510.00552"
}

@incollection{barnsteiner2008medication,
  author    = {Barnsteiner, J. H.},
  title     = {Medication Reconciliation},
  editor    = {Hughes, R. G.},
  booktitle = {Patient Safety and Quality: An Evidence-Based Handbook for Nurses},
  publisher = {Agency for Healthcare Research and Quality (US)},
  address   = {Rockville (MD)},
  year      = {2008},
  month     = {apr},
  chapter   = {38},
  url       = {https://www.ncbi.nlm.nih.gov/books/NBK2648/}
}

@inproceedings{Park2023-ACM,
author = {Park, Joon Sung and O'Brien, Joseph and Cai, Carrie Jun and Morris, Meredith Ringel and Liang, Percy and Bernstein, Michael S.},
title = {Generative Agents: Interactive Simulacra of Human Behavior},
year = {2023},
isbn = {9798400701320},
publisher = {Association for Computing Machinery},
address = {New York, NY, USA},
url = {https://doi.org/10.1145/3586183.3606763},
doi = {10.1145/3586183.3606763},
booktitle = {Proceedings of the 36th Annual ACM Symposium on User Interface Software and Technology},
articleno = {2},
numpages = {22},
location = {San Francisco, CA, USA},
series = {UIST '23}
}

@ARTICLE{Packer2023-gk,
  title         = "{MemGPT}: Towards {LLMs} as operating systems",
  author        = "Packer, Charles and Wooders, Sarah and Lin, Kevin and Fang,
                   Vivian and Patil, Shishir G and Stoica, Ion and Gonzalez,
                   Joseph E",
  month         =  oct,
  year          =  2023,
  copyright     = "http://creativecommons.org/licenses/by/4.0/",
  archivePrefix = "arXiv",
  primaryClass  = "cs.AI",
  eprint        = "2310.08560"
}

@ARTICLE{Chhikara2025-fb,
  title         = "Mem0: Building production-ready {AI} agents with scalable
                   long-term memory",
  author        = "Chhikara, Prateek and Khant, Dev and Aryan, Saket and Singh,
                   Taranjeet and Yadav, Deshraj",
  month         =  apr,
  year          =  2025,
  copyright     = "http://arxiv.org/licenses/nonexclusive-distrib/1.0/",
  archivePrefix = "arXiv",
  primaryClass  = "cs.CL",
  eprint        = "2504.19413"
}

@ARTICLE{Maharana2024-uc,
  title         = "Evaluating very long-term conversational memory of {LLM}
                   agents",
  author        = "Maharana, Adyasha and Lee, Dong-Ho and Tulyakov, Sergey and
                   Bansal, Mohit and Barbieri, Francesco and Fang, Yuwei",
  month         =  feb,
  year          =  2024,
  copyright     = "http://creativecommons.org/licenses/by-nc-sa/4.0/",
  archivePrefix = "arXiv",
  primaryClass  = "cs.CL",
  eprint        = "2402.17753"
}

@ARTICLE{Ai2025-kb,
  title         = "{MemoryBench}: A benchmark for memory and continual learning
                   in {LLM} systems",
  author        = "Ai, Qingyao and Tang, Yichen and Wang, Changyue and Long,
                   Jianming and Su, Weihang and Liu, Yiqun",
  month         =  dec,
  year          =  2025,
  copyright     = "http://arxiv.org/licenses/nonexclusive-distrib/1.0/",
  archivePrefix = "arXiv",
  primaryClass  = "cs.LG",
  eprint        = "2510.17281"
}

@ARTICLE{Wu2024-mo,
  title         = "{LongMemEval}: Benchmarking Chat Assistants on {Long-Term}
                   Interactive Memory",
  author        = "Wu, Di and Wang, Hongwei and Yu, Wenhao and Zhang, Yuwei and
                   Chang, Kai-Wei and Yu, Dong",
  month         =  oct,
  year          =  2024,
  copyright     = "http://creativecommons.org/licenses/by/4.0/",
  archivePrefix = "arXiv",
  primaryClass  = "cs.CL",
  eprint        = "2410.10813"
}

@ARTICLE{Lee2025-tu,
  title         = "{FHIR-AgentBench}: Benchmarking {LLM} Agents for Realistic
                   Interoperable {EHR} Question Answering",
  author        = "Lee, Gyubok and Bach, Elea and Yang, Eric and Pollard, Tom
                   and Johnson, Alistair and Choi, Edward and Jia, Yugang and
                   Lee, Jong Ha",
  month         =  nov,
  year          =  2025,
  copyright     = "http://creativecommons.org/licenses/by/4.0/",
  archivePrefix = "arXiv",
  primaryClass  = "cs.CL",
  eprint        = "2509.19319"
}

@article{Jiang-2025-mab,
author = {Yixing Jiang  and Kameron C. Black  and Gloria Geng  and Danny Park  and James Zou  and Andrew Y. Ng  and Jonathan H. Chen },
title = {MedAgentBench: A Virtual EHR Environment to Benchmark Medical LLM Agents},
journal = {NEJM AI},
volume = {2},
number = {9},
pages = {AIdbp2500144},
year = {2025},
doi = {10.1056/AIdbp2500144},

URL = {https://ai.nejm.org/doi/full/10.1056/AIdbp2500144},
eprint = {https://ai.nejm.org/doi/pdf/10.1056/AIdbp2500144}
}

@ARTICLE{Schmiedmayer2025-qp,
  title     = "{LLMonFHIR}: A physician-validated, large language model-based
               mobile application for querying patient electronic health data",
  author    = "Schmiedmayer, Paul and Rao, Adrit and Zagar, Philipp and Aalami,
               Lauren and Ravi, Vishnu and Zahedivash, Aydin and Yao, Dong-Han
               and Fereydooni, Arash and Aalami, Oliver",
  journal   = "JACC Adv.",
  publisher = "Elsevier BV",
  volume    =  4,
  number    = "6 Pt 1",
  pages     = "101780",
  month     =  jun,
  year      =  2025,
  copyright = "http://creativecommons.org/licenses/by/4.0/",
  language  = "en"
}

@ARTICLE{Qu2026-ky,
  title         = "{PrecLLM}: A privacy-preserving framework for efficient
                   clinical annotation extraction from unstructured {EHRs}
                   using small-scale {LLMs}",
  author        = "Qu, Yixiang and Dai, Yifan and Yu, Shilin and Tanikella,
                   Pradham and Pillai, Malvika and Chen, Walter and Xie, Jialiu
                   and Ren, Yishan and Wang, Duan and Wang, Yikai and Sheth,
                   Sid and Chen, Guanting and Liu, Yufeng and Schrank, Travis
                   and Hackman, Trevor and Li, Didong and Wu, Di",
  month         =  mar,
  year          =  2026,
  copyright     = "http://arxiv.org/licenses/nonexclusive-distrib/1.0/",
  archivePrefix = "arXiv",
  primaryClass  = "cs.AI",
  eprint        = "2412.02868"
}

@ARTICLE{Alsentzer2019-no,
  title         = "Publicly available clinical {BERT} embeddings",
  author        = "Alsentzer, Emily and Murphy, John R and Boag, Willie and
                   Weng, Wei-Hung and Jin, Di and Naumann, Tristan and
                   McDermott, Matthew B A",
  month         =  apr,
  year          =  2019,
  copyright     = "http://arxiv.org/licenses/nonexclusive-distrib/1.0/",
  archivePrefix = "arXiv",
  primaryClass  = "cs.CL",
  eprint        = "1904.03323"
}

@ARTICLE{Rasmy2021-yk,
  title     = "{Med-BERT}: pretrained contextualized embeddings on large-scale
               structured electronic health records for disease prediction",
  author    = "Rasmy, Laila and Xiang, Yang and Xie, Ziqian and Tao, Cui and
               Zhi, Degui",
  journal   = "NPJ Digit. Med.",
  publisher = "Springer Science and Business Media LLC",
  volume    =  4,
  number    =  1,
  pages     = "86",
  month     =  may,
  year      =  2021,
  copyright = "https://creativecommons.org/licenses/by/4.0",
  language  = "en"
}

@UNPUBLISHED{Mokssit2026-fj,
  title    = "{FHIR-AgentEval}: A modular sandbox for benchmarking clinical
              {LLM} agents with an evaluation of memory-augmented
              configurations",
  author   = "Mokssit, Youssef and Ravi, Kamalakkannan and Nie, Mengshu and
              Kim, Junyoung and Liu, Cong",
  journal  = "Res. Sq.",
  month    =  feb,
  year     =  2026,
  language = "en"
}

@UNPUBLISHED{Deo2025-lk,
  title     = "A conversational artificial intelligence agent for medication
               reconciliation and review",
  author    = "Deo, Rahul C and Goto, Shinichi and Jain, Tara and Meier, Sarah
               and Patel, Rahul",
  journal   = "medRxiv",
  month     =  jun,
  year      =  2025,
  copyright = "http://creativecommons.org/licenses/by-nd/4.0/"
}

@inproceedings{gupta-etal-2020-human,
    title = "Human-Human Health Coaching via Text Messages: Corpus, Annotation, and Analysis",
    author = "Gupta, Itika  and
      Di Eugenio, Barbara  and
      Ziebart, Brian  and
      Baiju, Aiswarya  and
      Liu, Bing  and
      Gerber, Ben  and
      Sharp, Lisa  and
      Nabulsi, Nadia  and
      Smart, Mary",
    booktitle = "Proceedings of the 21th Annual Meeting of the Special Interest Group on Discourse and Dialogue",
    month = jul,
    year = "2020",
    address = "1st virtual meeting",
    publisher = "Association for Computational Linguistics",
    url = "https://aclanthology.org/2020.sigdial-1.30",
    pages = "246--256",
}

@ARTICLE{Walonoski2018-ci,
  title     = "Synthea: An approach, method, and software mechanism for
               generating synthetic patients and the synthetic electronic
               health care record",
  author    = "Walonoski, Jason and Kramer, Mark and Nichols, Joseph and Quina,
               Andre and Moesel, Chris and Hall, Dylan and Duffett, Carlton and
               Dube, Kudakwashe and Gallagher, Thomas and McLachlan, Scott",
  journal   = "J. Am. Med. Inform. Assoc.",
  publisher = "Oxford University Press (OUP)",
  volume    =  25,
  number    =  3,
  pages     = "230--238",
  month     =  mar,
  year      =  2018,
  copyright = "http://creativecommons.org/licenses/by-nc/4.0/",
  language  = "en"
}

@ARTICLE{OpenAI2024-ac,
  title         = "{GPT-4o} System Card",
  author        = "{OpenAI}",
  month         =  oct,
  year          =  2024,
  copyright     = "http://creativecommons.org/licenses/by/4.0/",
  archivePrefix = "arXiv",
  primaryClass  = "cs.CL",
  eprint        = "2410.21276"
}

@ARTICLE{Rasmussen2025-zep,
    title         = "Zep: A Temporal Knowledge Graph Architecture for Agent Memory",
    author        = "Rasmussen, Preston and Paliychuk, Pavlo and Beauvais, Travis
                     and Ryan, Jack and Chalef, Daniel",
    month         = jan,
    year          = 2025,
    archivePrefix = "arXiv",
    primaryClass  = "cs.AI",
    eprint        = "2501.13956",
    url           = "https://arxiv.org/abs/2501.13956"
  }

@ARTICLE{Hu2025-mab,
    title         = "Evaluating Memory in {LLM} Agents via Incremental
                     Multi-Turn Interactions",
    author        = "Hu, Yuanzhe and Wang, Yu and McAuley, Julian",
    month         = jul,
    year          = 2025,
    note          = "Accepted at ICLR 2026",
    archivePrefix = "arXiv",
    primaryClass  = "cs.CL",
    eprint        = "2507.05257",
    url           = "https://arxiv.org/abs/2507.05257"
  }

@ARTICLE{Sridharan2024-medrec,
    title     = "Unlocking the potential of advanced large language models in
                 medication review and reconciliation",
    author    = "Sridharan, Kannan and Sivaramakrishnan, Gowri",
    journal   = "Explor. Res. Clin. Soc. Pharm.",
    volume    = 15,
    pages     = "100492",
    month     = sep,
    year      = 2024,
    doi       = "10.1016/j.rcsop.2024.100492"
  }

@ARTICLE{Myakala2026-bs,
    title         = "BeliefShift: Benchmarking Temporal Belief Consistency and
                     Opinion Drift in {LLM} Agents",
    author        = "Myakala, Praveen Kumar and others",
    month         = mar,
    year          = 2026,
    archivePrefix = "arXiv",
    eprint        = "2603.23848"
  }

@ARTICLE{Gwet2008-ac1,
    title   = "Computing inter-rater reliability and its variance in the
               presence of high agreement",
    author  = "Gwet, Kilem Li",
    journal = "Br. J. Math. Stat. Psychol.",
    volume  = 61,
    number  = 1,
    pages   = "29--48",
    year    = 2008,
    doi     = "10.1348/000711006X126600"
  }

\newpage
\appendix
\renewcommand{\thetable}{A.\arabic{table}}
\setcounter{table}{0}

\section{Qualitative Examples}
\label{app:qualitative}

\begin{table}[h]
    \centering
    \caption{Qualitative examples of end-to-end pipeline behavior. The success case shows a medication discrepancy correctly detected with precise FHIR citation. The failure case shows extraction losing safety-critical details, causing the engine to miss a documented allergy conflict.}
    \small
    \begin{tabular}{p{2.2cm}p{5.5cm}p{5.5cm}}
    \toprule
      & \textbf{Success: Medication Change} & \textbf{Failure: Allergy Conflict} \\
      \midrule
      \textbf{Patient \newline utterance}
      & ``I stopped taking that lisinopril a few days ago because I think it was making me dizzy.''
      & ``I saw some peanut butter ones that looked good for a quick boost. I'm not allergic to anything major, so it should be fine.'' \\
      \addlinespace
      \textbf{GT memory}
      & ``Patient stopped taking lisinopril because they believe it was causing dizziness.''
      & ``Patient is considering eating a peanut butter protein bar and reports having no major allergies.'' \\
      \addlinespace
      \textbf{Extracted memory}
      & ``Patient stopped taking lisinopril a few days ago due to dizziness (reported Aug 21, 2019).''
      & ``The patient is considering taking a snack, like a protein bar, during walks to help with energy levels.'' \\
      \addlinespace
      \textbf{FHIR record}
      & \texttt{MedicationRequest}: lisinopril 10\,MG (active)
      & \texttt{AllergyIntolerance}: Allergy to peanuts (active) \\
      \addlinespace
      \textbf{Engine \newline result}
      & \texttt{contradiction}, severity: high
      & \texttt{no\_fhir} (routine snack preference) \\
      \addlinespace
      \textbf{Analysis}
      & Extraction preserved the clinical detail. Engine correctly identified the conflict with the active prescription.
      & Extraction dropped both ``peanut butter'' and the allergy denial (``not allergic to anything major''), reducing the memory to a generic snack preference. Without those details, the engine could not identify the conflict with the documented peanut allergy. \\
    \bottomrule
    \end{tabular}
    \label{tab:qualitative}
\end{table}

\section{Sample Clinical Summary}
\label{app:clinical_summary}

The following is an excerpt of the curated clinical summary provided to the reconciliation engine for one patient. Each resource includes inline code tags (e.g., \texttt{[SNOMED-CT:59621000]}) enabling the LLM to cite specific FHIR resources in its structured output. The full summary for this patient contains 33 conditions, 9 medications, 45 procedures, and 6 care plan activities across approximately 1,100 tokens. The excerpt below shows only a representative subset.

\begin{small}
\begin{verbatim}
CONDITIONS:
  - [Condition] Essential hypertension (disorder)
    (active) [onset: 1996-02-20] [SNOMED-CT:59621000]
  - [Condition] Prediabetes
    [onset: 1999-03-09] [SNOMED-CT:15777000]
  - [Condition] Anemia (disorder)
    [onset: 2000-03-14] [SNOMED-CT:271737000]
  - [Condition] Chronic sinusitis (disorder)
    [onset: 2009-09-04] [SNOMED-CT:40055000]
  - [Condition] Body mass index 30+ - obesity (finding)
    [onset: 2015-06-09] [SNOMED-CT:162864005]
  - [Condition] Ischemic heart disease (disorder)
    [onset: 2018-06-26] [SNOMED-CT:414545008]

MEDICATIONS:
  - [MedicationRequest] lisinopril 10 MG Oral Tablet
    (active) [authored: 2019-07-02] [RxNorm:314076]
  - [MedicationRequest] Clopidogrel 75 MG Oral Tablet
    (active) [authored: 2018-07-03] [RxNorm:309362]
  - [MedicationRequest] Simvastatin 20 MG Oral Tablet
    (active) [authored: 2018-07-03] [RxNorm:312961]
  - [MedicationRequest] Amoxicillin 250 MG Oral Capsule
    (stopped) [authored: 2010-04-19] [RxNorm:308182]

CARE PLANS:
  - [CarePlan] Prediabetes: Diabetic diet
    [SNOMED-CT:160670007]
  - [CarePlan] Prediabetes: Exercise therapy
    [SNOMED-CT:229065009]
  - [CarePlan] Essential hypertension: Prescribed
    activity/exercise education [SNOMED-CT:386463000]

KEY OBSERVATIONS (most recent):
  - [Observation] Body mass index (BMI) [Ratio]
    = 30.44 kg/m2 [2019-07-02] [LOINC:39156-5]
  - [Observation] Blood Pressure panel
    = Systolic: 107.0 mmHg / Diastolic: 73.0 mmHg
    [2019-07-02] [LOINC:85354-9]
  - [Observation] Hemoglobin A1c/Hemoglobin.total
    = 5.89 % [2019-07-02] [LOINC:4548-4]
\end{verbatim}
\end{small}

\section{Sample Reconciliation Result}
\label{app:recon_result}

The following is an actual engine output for a patient who mentions expecting an Amoxicillin prescription while claiming to have no allergies. The engine identifies the relevant FHIR resources (chronic sinusitis condition and stopped Amoxicillin prescription) and flags the clinical gap.

\begin{small}
\begin{verbatim}
Input memory: "Patient reports a recurring sinus issue
  is acting up, has a doctor's appointment for it
  tomorrow, and expects to be prescribed Amoxicillin,
  noting they have no allergies."
Category: health

Output:
{
  "contradiction_type": "gap_patient",
  "confidence": 0.95,
  "clinical_severity": "medium",
  "justification": "The patient's memory indicates a
    recurring sinus issue and expectation of being
    prescribed Amoxicillin, which is not currently
    active in the clinical record. The clinical record
    shows a history of chronic sinusitis but does not
    reflect the current recurrence or the anticipated
    prescription of Amoxicillin. This discrepancy is
    clinically actionable as it involves a potential
    change in medication management.",
  "fhir_resources_considered": [
    {
      "resource_type": "Condition",
      "code_system": "SNOMED-CT",
      "code_value": "40055000",
      "display": "Chronic sinusitis (disorder)"
    },
    {
      "resource_type": "MedicationRequest",
      "code_system": "RxNorm",
      "code_value": "308182",
      "display": "Amoxicillin 250 MG Oral Capsule
        (stopped)"
    }
  ]
}
\end{verbatim}
\end{small}

\section{MemoryDelta Schema}
\label{app:memory_schema}

The extraction LLM returns a structured output conforming to the following schema:

\begin{small}
\begin{verbatim}
class MemoryInsert:
    content: str     # Memory content to store
    category: str    # preference|health|lifestyle|
                     # medication|fact

class MemoryUpdate:
    memory_id: str   # ID of existing memory to update
    new_content: str # Replacement content
    category: str

class MemoryDelete:
    memory_id: str   # ID of existing memory to remove
    justification: str

class MemoryDelta:
    inserts: List[MemoryInsert]
    updates: List[MemoryUpdate]
    deletes: List[MemoryDelete]
\end{verbatim}
\end{small}

\section{Reconciliation Output Schema}
\label{app:recon_schema}

\begin{small}
\begin{verbatim}
class FHIRResourceRef:
    resource_type: str  # e.g., "MedicationRequest"
    code_system: str    # e.g., "RxNorm"
    code_value: str     # e.g., "314076"
    display: str        # e.g., "lisinopril 10 MG"

class ReconciliationResult:
    contradiction_type: str  # agreement|contradiction|
                             # gap_patient|no_fhir
    confidence: float        # 0.0-1.0
    justification: str
    clinical_severity: str   # low|medium|high
    fhir_resources_considered: List[FHIRResourceRef]
\end{verbatim}
\end{small}

\section{Human Annotation Instructions}
\label{app:human_annotation}

Two digital health practitioners with clinical backgrounds independently annotated extracted memory events for a subset of patients. The following instructions were provided verbatim:

\begin{small}
\begin{verbatim}
MEMORY EXTRACTION ANNOTATION

Thank you for participating in this evaluation. You are
reviewing an AI system that extracts "memory events" from
health coaching conversations — facts, preferences, and
events that a coach should remember about a patient for
future sessions.

WHAT YOU'LL SEE

Each patient has their own tab. Within each tab:

  - Session transcripts: These come from real text-message
    coaching conversations conducted over several weeks.
    The original data is one long thread of messages; we
    segmented it into sessions based on gaps in the
    conversation. Sessions labeled "UIC" are these real
    coaching conversations. Sessions labeled "Clinical"
    are simulated clinical check-in sessions we created
    to test the system's ability to capture health-related
    information (e.g., medication changes, symptoms).
    Read all sessions in order, top to bottom.

  - Memory events: After each session's transcript, you'll
    see a list of facts the AI extracted. Your job is to
    rate each one.

  - Missing memories: After each session's events, there's
    a prompt to note anything the AI missed.

HOW TO ANNOTATE

For each memory event (white rows with an Event ID),
select a rating from the dropdown in two columns:

ACCURACY — Does this memory event correctly reflect what
the patient communicated?

  Accurate:
    The event faithfully and completely captures what the
    patient said. The meaning is preserved without
    distortion, exaggeration, or omission of key context.

  Partially Accurate:
    The core idea is correct, but there is a meaningful
    issue:
    - Important qualifying detail is missing (e.g.,
      "takes medication" but omits "only on weekdays")
    - Timeframe or frequency is slightly off
    - Phrasing implies more certainty than the patient
      expressed

  Inaccurate:
    The event misrepresents what the patient said,
    attributes something they did not say, or gets a key
    detail wrong (e.g., wrong medication name, wrong
    date, opposite sentiment).

RELEVANCE — Would a health coach benefit from remembering
this for future sessions?

  Essential:
    A coach needs this to provide safe, effective care.
    Examples:
    - Current goals, targets, and confidence levels
    - Health conditions, symptoms, or physical limitations
      affecting activity
    - Medications (current, stopped, or changed)
    - Major barriers to progress

  Useful:
    Helpful context that enriches coaching but isn't
    critical. Examples:
    - Preferred walking times or locations
    - Work schedule or lifestyle context
    - One-time events (travel, family events) that
      affected a specific week
    - Patient preferences or communication style

  Not Useful:
    Not worth storing as a persistent memory. Examples:
    - Trivial pleasantries or acknowledgments
    - Information already obvious from context
    - Overly specific details unlikely to matter later
      (e.g., exact step count on one day)

NOTES (per event)

Use the Notes column on each memory event row to add any
clarification — e.g., why you rated it Partially
Accurate, or additional context the event should have
included.

MISSING MEMORIES & SESSION NOTES

After each session's events, you'll see two prompt rows:
  - "Anything important from Session X NOT captured
    above?" — type your response in the merged cell to
    the right of the prompt.
  - "General notes on Session X?" — same thing, type in
    the merged cell to the right.

Focus on things a coach would want to remember — don't
worry about trivial details.

IMPORTANT NOTES
  - Sessions labeled "Clinical" are simulated clinical
    check-ins (not real conversations). Please still
    evaluate the memory events from these sessions.
  - Read sessions in order — context from earlier
    sessions matters.
  - There are no right or wrong answers. We want your
    professional judgment.
\end{verbatim}
\end{small}

\section{Memory Extraction Prompt}
\label{app:extraction_prompts}

The following prompt is provided as system instructions to the extraction LLM (GPT-4o) along with the current memory state formatted with inline IDs. The LLM returns a structured \texttt{MemoryDelta} object containing explicit insert, update, and delete operations.

\begin{small}
\begin{verbatim}
Extract important information from the conversation to
remember for future sessions.

=== CORE PRINCIPLE: STABLE FACTS vs. EVOLVING STATE ===

Every piece of information falls into one of three types.
The type determines whether it gets its own memory or is
consolidated with others.

**Stable facts** — things about the patient that remain
true regardless of how their goals change. Each stable
fact gets its OWN separate memory. NEVER fold a stable
fact into a larger evolving-state memory, because it will
be lost when that memory updates.
Examples of stable fact categories (DO NOT use these
specific examples as extracted memories — only extract
what the patient actually says):
- Equipment owned (e.g., exercise equipment at home)
- Job or daily routine context (e.g., active job that
  contributes to step count)
- Exercise habits or repertoire beyond the current goal
- Preferences (e.g., communication preferences, coaching
  style preferences)
- Preferred locations for activity
- Coping strategies or backup plans

**Evolving state** — things that change over time. These
belong in a single memory per topic that gets UPDATED as
the state changes. Include brief progression history.
Examples of evolving state:
- Current goal, target days, confidence level, and active
  barriers — all in one memory, updated each time any of
  these change
- When the goal changes, update the existing memory to
  reflect the new goal while noting the progression

**Significant events** — notable things that happened at
a specific point in time. Each gets its own memory with
the date.
Examples of significant events:
- Illness or injury that affected the program
- Travel or time away
- Life events that disrupted routine

IMPORTANT: Only extract information the patient actually
stated in the conversation. NEVER insert information from
these instructions or examples.

**Key test**: If the patient's current goal changes next
week, would this fact STILL be true? If yes -> stable
fact -> separate memory. If no -> evolving state ->
update the existing goal memory.

=== DATES ON EVERYTHING ===

Every memory must include a date indicating when the
information was first reported or last changed. The
session date is provided — use it.

- Stable facts: include the date first mentioned,
  e.g., "(reported [session date])"
- Evolving state: include the date of the latest change,
  e.g., "as of [session date] (changed from X)"
- Events: include when it happened,
  e.g., "[description] on/around [session date]"
- NEVER use relative time ("last week", "yesterday")
  without an absolute date.

=== WHEN TO UPDATE vs. INSERT vs. DELETE vs. NO CHANGE ===

- Evolving state changed -> add an UPDATE with the
  existing memory's memory_id
- New stable fact or new event -> add an INSERT
- Two existing memories overlap on the same evolving
  state -> UPDATE one, DELETE the other
- Nothing new -> return empty lists for all three
  operations

UPDATE and DELETE operations MUST reference a valid
memory_id from the existing memories list. Do NOT
fabricate IDs. If you want to modify a memory but cannot
find its ID, use INSERT instead.

=== DUPLICATE PREVENTION ===

Before adding an INSERT, carefully scan the EXISTING
MEMORIES list above. If an existing memory already
captures the same fact, preference, or event — even if
worded slightly differently — do NOT insert a duplicate.
Instead:
- If the existing memory is accurate and complete ->
  do nothing (return empty lists)
- If the existing memory needs updating -> use UPDATE
  with its memory_id

Never insert a new memory for evolving state that an
existing memory already tracks. But DO insert separate
memories for new stable facts, even if they are
thematically related to an existing evolving-state memory.

=== EVOLVING STATE GRANULARITY ===

For evolving state, group closely-coupled fields into one
memory. Keep unrelated state separate.

- GOOD: One memory for "current step goal + target days +
  confidence + active barriers" — these change together
  and form a coherent program snapshot.
- BAD: Separate memories for "step goal is 5,000" and
  "goal days are Mon/Wed/Thu" and "confidence is 10/10"
  — these are tightly coupled state about the same
  program.
- BAD: One memory combining "step goal state" and
  "dietary goal state" — these are independent evolving
  states.

Summarize trends rather than listing individual data
points: "consistently exceeds goal, typically 6K-9K
daily steps" NOT each day's count.

=== FAITHFULNESS ===

- Preserve the user's original meaning. Record what they
  said, not a reinterpretation.
- Aspirations and intentions must stay distinguishable
  from current behavior. "I want to do X" != "does X".
- Only extract information explicitly stated — do not
  infer.

=== CATEGORIES ===

Assign one category per memory (for organization only):
- **preference**: Name/nickname, communication style,
  language preference
- **health**: Symptoms, conditions, physical limitations
- **lifestyle**: Diet, exercise, work schedule, sleep
  patterns, habits
- **medication**: OTC medications, supplements
- **fact**: Other important facts

=== NO CHANGES ===

If the conversation contains NO new information worth
remembering, return empty lists for inserts, updates,
and deletes.
\end{verbatim}
\end{small}

\section{Reconciliation Engine Prompt}
\label{app:recon_prompt}

The following prompt is provided to the reconciliation LLM (GPT-4o) along with the patient's extracted memory and the curated clinical summary. The memory content, category, and full clinical summary are inserted into the template at runtime.

\begin{small}
\begin{verbatim}
You are a clinical reconciliation engine. Given a
patient's memory from a health coaching conversation and
their clinical record summary, determine whether there is
a clinically relevant discrepancy.

## Patient Memory
Content: "{content}"
Category: {category}

## Clinical Record Summary
{clinical_summary}

## Classification Types

- **contradiction**: The memory directly conflicts with
  the clinical record. Example: patient denies a
  documented allergy, claims they stopped a medication
  that is still active, or denies a diagnosed condition.

- **gap_patient**: The memory contains a specific
  clinical assertion that is absent from the clinical
  record AND is clinically actionable. This includes:
  undocumented medications or supplements the patient is
  taking, unreported symptoms, self-reported condition
  changes (e.g., "my anemia resolved years ago" when
  FHIR still shows active), or a specific activity plan
  that poses safety risk given their documented
  conditions (e.g., planning HIIT with active cardiac
  disease).

- **agreement**: The memory is consistent with the
  clinical record. The patient's statement aligns with
  or confirms what is documented.

- **no_fhir**: The memory has no specific clinical
  relevance. This is the correct classification for
  routine coaching content: step goals, walking targets,
  exercise schedules, daily routines, confidence levels,
  preferences, communication style, and general lifestyle
  facts. These should be classified as no_fhir even if
  the patient has conditions broadly related to exercise
  or diet (e.g., a step goal is no_fhir even if the
  patient has prediabetes with an exercise therapy care
  plan — the patient is following their plan, not
  conflicting with it). A memory is only clinically
  relevant if it makes a specific assertion about
  medications, symptoms, diagnoses, allergies, a concrete
  activity that is contraindicated by their conditions,
  or a specific barrier to following a named clinical
  recommendation (e.g., "can't follow the DASH diet due
  to work schedule" references a specific care plan and
  is gap_patient, but "too busy to walk today" is
  no_fhir).

## Instructions
1. First, determine if the memory has specific clinical
   relevance. If it describes routine coaching activity
   (goals, steps, preferences, schedules, barriers),
   classify as no_fhir and stop — do not search for
   tenuous connections to care plans or conditions.
2. If clinically relevant, compare against the clinical
   summary.
3. Consider data freshness — older clinical records may
   be stale.
4. Assess clinical severity: low, medium, high.
5. For each relevant clinical resource, provide its
   resource_type, code_system, code_value, and display
   exactly as shown in the clinical summary brackets
   [SYSTEM:CODE].
6. Provide a confidence score (0.0-1.0) and brief
   justification.
\end{verbatim}
\end{small}

\section{LLM-as-Judge Prompts}
\label{app:judge_prompts}

\subsection{GT Event Matching}

For each ground-truth memory event, an LLM judge (GPT-4o) evaluates whether the information is captured in the system's memory state at the corresponding point in the conversation:

\begin{small}
\begin{verbatim}
You are evaluating whether a memory extraction system
captured a specific piece of information from a coaching
conversation.

## Expected Information (Ground Truth)
The patient said: "{utterance}"
This should have been remembered as: "{memory_content}"

## Memory State After Processing That Utterance
These are ALL memories the system had at that point in
time:

{memories}

## Task
Determine if the expected information is captured in the
memory state. The information may be spread across
multiple memories — that is fine. Look across ALL
memories in combination, not just a single one.

- MATCH: The expected information is fully captured
  across the memory state (may use different words, may
  be split across multiple memories)
- PARTIAL: Only some of the expected information appears
  in the memories (key details are missing entirely, not
  just rephrased)
- NO_MATCH: None of the expected information appears in
  any memory

Respond with valid JSON only, no markdown:
{
  "verdict": "MATCH|PARTIAL|NO_MATCH",
  "content_fidelity": null or 1-5
    (5=perfect capture, 1=barely related),
  "reasoning": "brief explanation"
}
\end{verbatim}
\end{small}

\subsection{Transcript Judge}

A separate judge (GPT-4.1) evaluates the final memory state holistically against the complete conversation transcript across all sessions:

\begin{small}
\begin{verbatim}
You are evaluating a memory extraction system for a
health coaching application. You will read the full
conversation transcript between a coach and patient,
then evaluate the final memory state the system produced.

## Conversation Transcript
{transcript}

## Final Memory State
{memories}

## Task
Score the memory extraction on these 2 dimensions
(1-5 each):

1. **Faithfulness** (1-5): Are the memories faithful to
   what was actually said? No hallucinated, fabricated,
   or distorted information?
   - 5: All memories accurately reflect the conversation,
     no hallucinations
   - 3: Mostly accurate, some minor distortions or
     unsupported inferences
   - 1: Significant inaccuracies or hallucinated
     information

2. **Deduplication** (1-5): Is there one memory per
   concept? No redundant or overlapping memories?
   - 5: No duplicates, each memory is unique
   - 3: Some overlap but mostly distinct
   - 1: Heavy duplication, same info repeated

Respond with valid JSON only, no markdown:
{
  "faithfulness": { "score": 1-5, "reasoning": "..." },
  "deduplication": { "score": 1-5, "reasoning": "..." },
  "overall_notes": "any additional observations"
}
\end{verbatim}
\end{small}

\section*{LLM Usage Disclosure}

Large language models were used to assist with code development and manuscript editing during the preparation of this work. All LLM-generated content was reviewed and verified by the authors.
\end{document}